\documentclass{article}

\usepackage{microtype}
\usepackage{graphicx}
\usepackage{booktabs}
\usepackage{hyperref}
\usepackage[accepted]{icml2026custom}
\usepackage{amsmath,amssymb}

\icmltitlerunning{The Discrete-Log Clock}

\begin{document}

\twocolumn[
\icmltitle{The Discrete-Log Clock: How a Transformer Learns Modular Multiplication}

\begin{icmlauthorlist}
\icmlauthor{Huu Danh Nguyen}{stanford}
\end{icmlauthorlist}

\icmlaffiliation{stanford}{Stanford University}
\icmlcorrespondingauthor{Huu Danh Nguyen}{dannycodeinc@gmail.com}

\vskip 0.3in
]

\printAffiliationsAndNotice{}

\begin{abstract}
When small transformers grok modular multiplication, prior work reports that the learned embedding has a ``dense'' Fourier spectrum requiring all frequencies.
This contrasts with modular addition, where only a sparse set of key frequencies suffices.
We show this density is an artifact of analyzing in the wrong basis.
The natural Fourier transform for multiplication is not the standard additive DFT but the \emph{multiplicative character transform}, which decomposes functions on the multiplicative group $(\mathbb{Z}/p\mathbb{Z})^*$ into its irreducible representations.
Applying this transform to a grokked transformer trained on $a \cdot b \bmod 113$, we find the embedding spectrum becomes highly sparse (Gini coefficient 0.58 vs.\ 0.07 in the additive basis) with only 4 key frequencies carrying significant energy.
Furthermore, 96.9\% of MLP neurons are cleanly tuned to a single multiplicative frequency, and neuron activation heatmaps reveal 2D-periodic structure when reordered by the discrete logarithm.
These results demonstrate the transformer reduces multiplication to addition in discrete-log space, implementing a ``Discrete-Log Clock'' algorithm analogous to Nanda et al.'s Clock algorithm for addition.
The methodology generalizes: matching the analysis basis to the algebraic structure of the task reveals interpretable structure where standard tools see noise.
\end{abstract}

\section{Introduction}

Grokking, the phenomenon where neural networks suddenly generalize long after memorizing training data, has become a key testbed for mechanistic interpretability \citep{power2022grokking}.
For modular addition ($a + b \bmod p$), the internal algorithm has been fully reverse-engineered: the model learns a sparse Fourier representation and applies trigonometric identities to compose rotations on a circle \citep{nanda2023progress, zhong2023clock}.

For modular multiplication ($a \cdot b \bmod p$), models also grok successfully, but the internal algorithm has remained poorly characterized.
\citet{doshi2024grokking} derived analytical MLP solutions requiring dense Fourier components (all frequencies), and \citet{furuta2024interpreting} confirmed experimentally that grokked transformers use cosine-biased components across all frequencies.
These findings suggest multiplication uses a fundamentally different, less interpretable algorithm than addition.

\textbf{Our contribution.}
We show this conclusion is premature.
The ``dense spectrum'' is an artifact of analyzing with the \emph{additive} Fourier transform, which is the natural basis for addition but not for multiplication.
The correct analysis tool is the \emph{multiplicative character transform}, which is the Fourier transform on the multiplicative group $(\mathbb{Z}/p\mathbb{Z})^*$.
In this basis, the embedding becomes sparse, the neurons cluster cleanly by frequency, and the algorithm is interpretable: It reduces multiplication to addition via the discrete logarithm.

\textbf{Task formulation.}
The input to our model is a pair of integers $(a, b) \in \{1, \ldots, 112\}^2$.
We train a 1-layer transformer to output the predicted product $c = a \cdot b \bmod 113$.
The model observes 30\% of all input pairs during training and must generalize to the remaining 70\%.

\section{Related Work}

\textbf{Mechanistic interpretability of modular addition.}
\citet{nanda2023progress} fully reverse-engineered the algorithm learned by transformers on modular addition, showing the model uses discrete Fourier transforms and trigonometric identities to compose rotations.
\citet{zhong2023clock} named this the ``Clock'' algorithm and discovered an alternative ``Pizza'' algorithm, demonstrating that multiple algorithmic solutions exist for the same task.
These works establish the methodology we extend to multiplication.

\textbf{Prior work on modular multiplication.}
\citet{doshi2024grokking} derived analytical closed-form MLP weights for modular multiplication and found that the solutions contain the discrete logarithm $\log_g(i)$ in the weight structure.
However, they used quadratic activations and did not connect this to the empirical Fourier picture in trained ReLU transformers.
\citet{furuta2024interpreting} trained transformers on modular polynomials (including multiplication) and reported that the learned spectrum uses ``all frequencies'' with no clear sparse structure, in contrast to addition.
Our work resolves this discrepancy: the density is a basis artifact, not an algorithmic difference.

\textbf{Group-theoretic frameworks.}
\citet{chughtai2023toy} showed that networks trained on group operations learn irreducible representations of the relevant group.
\citet{stander2024grokking} reverse-engineered networks on permutation group multiplication (S5, S6), finding coset-based circuits.
\citet{mccracken2025universal} unified all known addition algorithms under an ``approximate CRT'' framework.
These works study different groups (non-abelian permutation groups, additive cyclic groups) from ours; our contribution is applying the multiplicative character basis specifically to $(\mathbb{Z}/p\mathbb{Z})^*$ in a trained transformer.

\section{Methods}

\subsection{Task and Architecture}

We train a 1-layer decoder-only transformer on $a \cdot b \bmod 113$ for all $a, b \in \{1, \ldots, 112\}$ (excluding zero, which lies outside the multiplicative group).
The architecture follows \citet{nanda2023progress}: $d_\text{model} = 128$, 4 attention heads ($d_\text{head} = 32$), MLP hidden dimension 512 with ReLU activation, no LayerNorm.
Input format is the token sequence $[a, b, {=}]$; the model's prediction is read from the output logits at position 2 (``${=}$'').

\textbf{Hyperparameters.}
We use 30\% training data (3,763 of 12,544 pairs), AdamW with learning rate $10^{-3}$, weight decay 1.0, and train for 40{,}000 epochs with full-batch gradient descent.
These hyperparameters exactly follow \citet{nanda2023progress} to ensure comparability; the only change is the task (multiplication vs.\ addition).

\begin{figure}[t]
\centering
\includegraphics[width=\columnwidth]{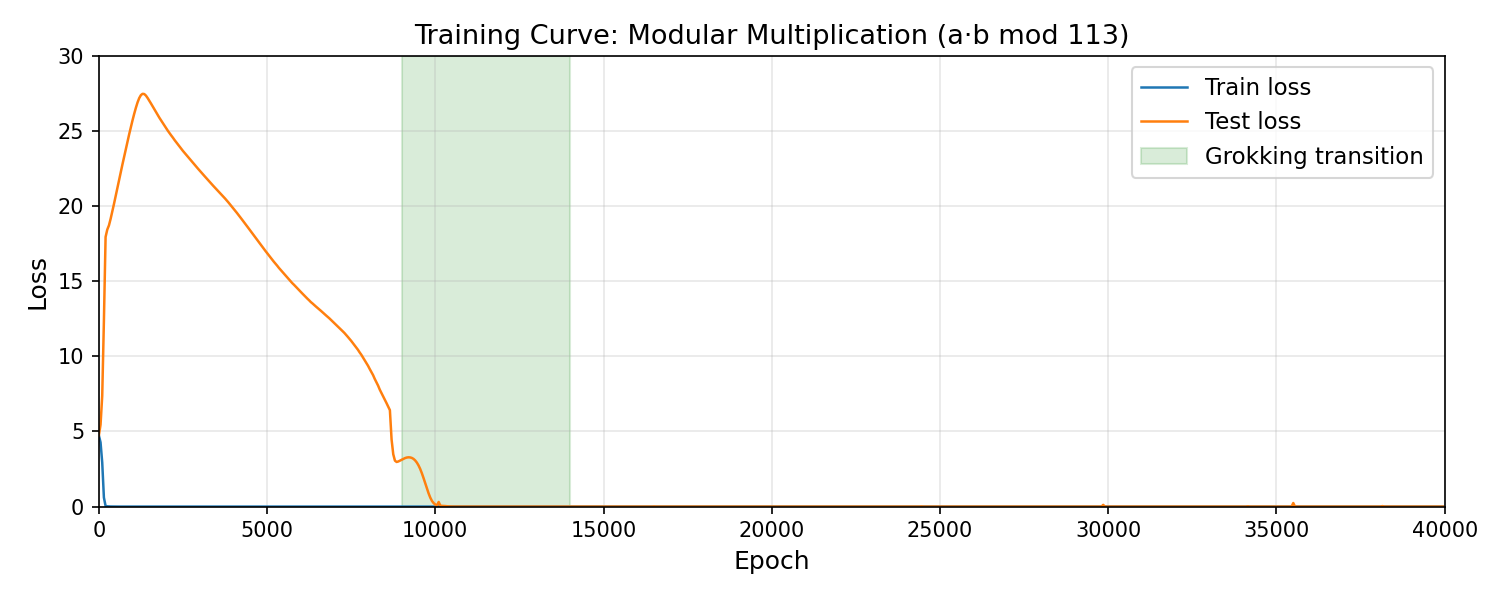}
\caption{Training curve for $a \cdot b \bmod 113$.
The model memorizes quickly (train loss drops by epoch 500), then generalizes suddenly during grokking (epochs 9K--14K).
Weight decay drives the transition from memorization to the structured algorithm.}
\label{fig:training}
\end{figure}

\textbf{Training dynamics.}
The model memorizes the training set by epoch $\sim$500 (train loss $\to 0$, test loss remains high).
Between epochs 9{,}000 and 14{,}000, test loss drops suddenly to near zero: the model groks, achieving near-perfect generalization (Figure~\ref{fig:training}).
This delayed generalization is characteristic of grokking: the model first overfits, then discovers the generalizing algorithm under pressure from weight decay.

\subsection{The Multiplicative Group and Discrete Logarithm}

Since 113 is prime, the nonzero integers $\{1, \ldots, 112\}$ form a cyclic group under multiplication mod 113.
A \emph{primitive root} is a generator: an element $g$ such that $\{g^0, g^1, \ldots, g^{111}\} \bmod 113$ exhausts all 112 elements.
For $p = 113$, we use $g = 3$.

The \emph{discrete logarithm} $\log_g\colon \{1, \ldots, 112\} \to \{0, \ldots, 111\}$ assigns each element its exponent: $3^\alpha \equiv a \pmod{113}$ means $\log_3(a) = \alpha$.
We write $\alpha = \log_g(a)$ and $\beta = \log_g(b)$ for the two inputs.
This is an exact bijection (a permutation of the 112 elements) that defines a group isomorphism:
\begin{equation}
\log_g : (\mathbb{Z}/p\mathbb{Z})^* \xrightarrow{\;\cong\;} \mathbb{Z}/(p{-}1)\mathbb{Z}
\end{equation}
Under this map, multiplication becomes addition:
\begin{equation}
a \cdot b \equiv 3^{(\alpha + \beta) \bmod 112} \pmod{113}
\label{eq:log_isomorphism}
\end{equation}
In the relabeled coordinates, the multiplication table \emph{is} the addition table mod 112.

\subsection{Two Fourier Bases}

The \textbf{additive} Fourier basis consists of $\sin(2\pi k a / q)$ and $\cos(2\pi k a / q)$ for $k = 1, \ldots, q/2$, where $q = 112$.
This is the standard DFT, natural for functions periodic in the integer label $a$.

The \textbf{multiplicative character basis} uses $\sin(2\pi k \log_g(a) / q)$ and $\cos(2\pi k \log_g(a) / q)$.
Operationally: reorder the embedding rows by the discrete-log map, then take the standard DFT.
This is the Fourier transform on the group $(\mathbb{Z}/p\mathbb{Z})^*$, decomposing functions into multiplicative characters.

The motivation is direct: since multiplication becomes addition after relabeling (Eq.~\ref{eq:log_isomorphism}), a model that has learned the group structure should have embeddings periodic in the relabeled coordinates, just as addition models have embeddings periodic in the original coordinates.

\subsection{Analysis Methodology}

We apply the following pipeline to the trained model:

\textbf{Step 1: Embedding spectrum.}
Extract the trained embedding matrix $W_E \in \mathbb{R}^{112 \times 128}$.
Project onto both bases and compute the combined frequency norm $\|f_k\| = \sqrt{\|s_k\|^2 + \|c_k\|^2}$ for each frequency $k$, where $s_k$ and $c_k$ are the sin and cos projections (each a 128-dim vector).

\textbf{Step 2: Sparsity metrics.}
We measure concentration using the Gini coefficient:
\begin{equation}
G = \frac{\sum_{i=1}^{n}(2i - n - 1)|x_i|}{\,n \sum_{i=1}^{n} |x_i|}
\end{equation}
where $x_1 \leq \ldots \leq x_n$ are the sorted frequency norms.
$G = 0$ means uniform (dense), $G = 1$ means maximally sparse.
Key frequencies are detected as those exceeding $5\times$ the median norm.

\textbf{Step 3: Neuron frequency assignment.}
For each of 512 MLP neurons, compute the 2D activation pattern $h_n(a,b)$ over all input pairs (reshaped to a $112 \times 112$ grid).
Take the 2D DFT in each basis and measure the maximum fraction of energy at any single frequency.
Neurons with $>$85\% energy at one frequency are classified as ``single-frequency tuned.''

\textbf{Step 4: SVD analysis.}
Perform SVD on $W_E$ and reorder the principal components by discrete logarithm.
Sinusoidal structure in log-order confirms the embedding encodes multiplicative characters.

\section{Experiments and Results}

\subsection{Basis Comparison}

\begin{figure}[t]
\centering
\includegraphics[width=\columnwidth]{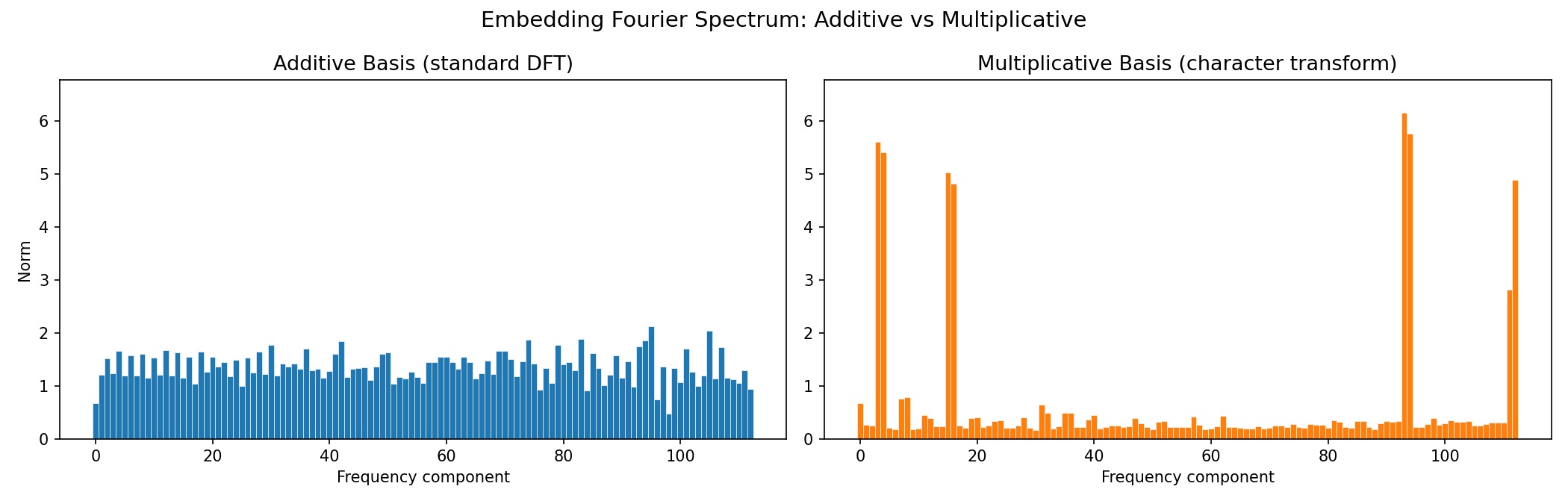}
\caption{Embedding Fourier spectrum.
\textbf{Left:} additive basis shows flat, dense spectrum.
\textbf{Right:} multiplicative basis reveals 4 sparse peaks.
Same y-axis scale.
The ``density'' reported by prior work is a basis artifact.}
\label{fig:spectrum}
\end{figure}

Figure~\ref{fig:spectrum} shows the embedding spectrum in both bases.
In the additive basis, energy spreads uniformly (Gini = 0.071).
In the multiplicative basis, energy concentrates in 4 peaks at frequencies $\{2, 8, 47, 56\}$ (Gini = 0.579, an 8.1$\times$ increase).
Table~\ref{tab:metrics} summarizes all metrics.

\begin{table}[t]
\centering
\caption{Sparsity metrics comparing the two bases.}
\label{tab:metrics}
\begin{tabular}{lcc}
\toprule
Metric & Additive & Multiplicative \\
\midrule
Gini coefficient & 0.071 & \textbf{0.579} \\
Inverse Participation Ratio & 52.7 & \textbf{4.1} \\
Key frequencies detected & 0 & \textbf{4} \\
Neurons $>$85\% single-freq & 0/512 & \textbf{496/512} \\
Mean max-fraction & 0.054 & \textbf{0.925} \\
\bottomrule
\end{tabular}
\end{table}

\subsection{Neuron-Level Confirmation}

\begin{figure}[t]
\centering
\includegraphics[width=\columnwidth]{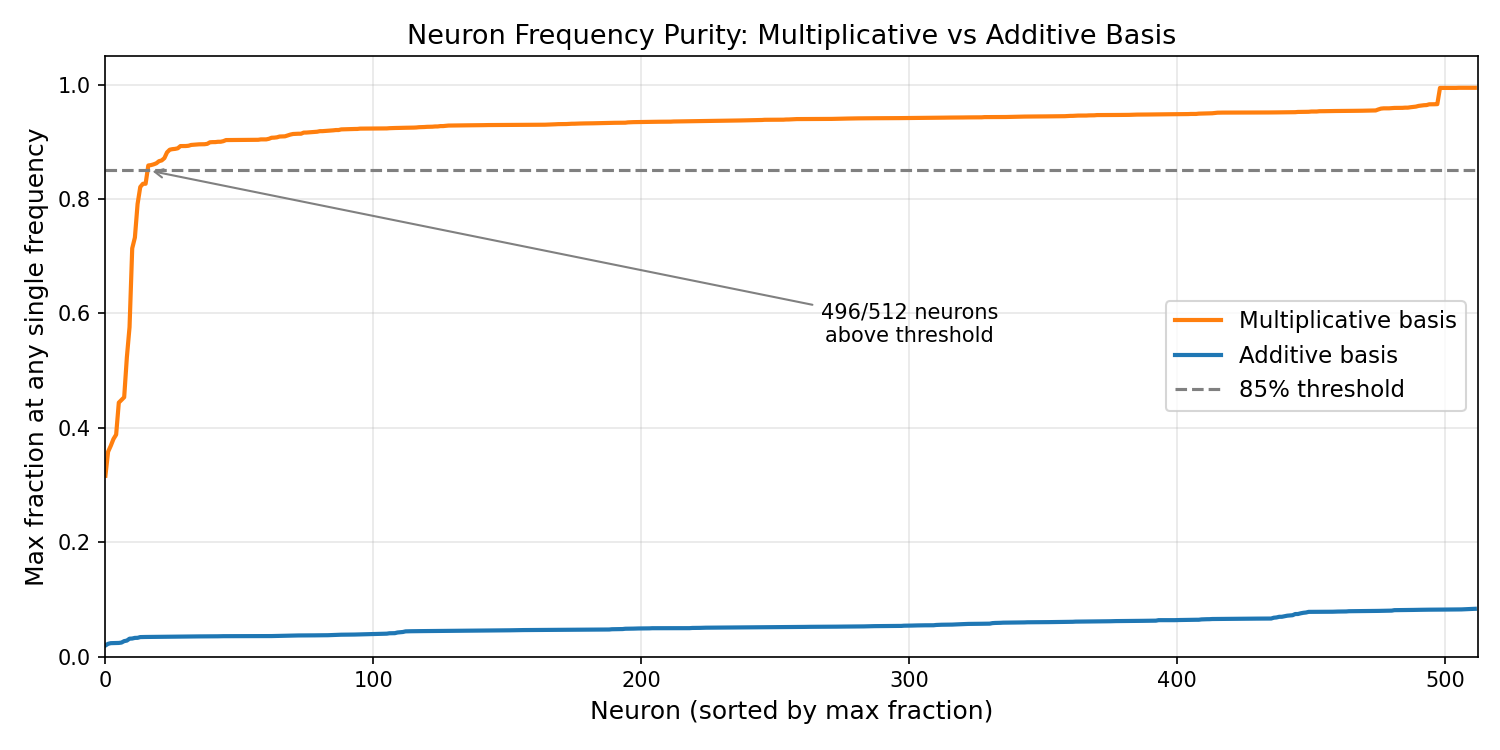}
\caption{Sorted maximum frequency fraction per neuron.
Orange: multiplicative basis (96.9\% above 85\% threshold).
Blue: additive basis (0\% above threshold).}
\label{fig:clustering}
\end{figure}

In the multiplicative basis, 496 out of 512 neurons (96.9\%) have $>$85\% of their Fourier energy at a single frequency (Figure~\ref{fig:clustering}).
In the additive basis, zero neurons pass this threshold.

\begin{figure}[t]
\centering
\includegraphics[width=\columnwidth]{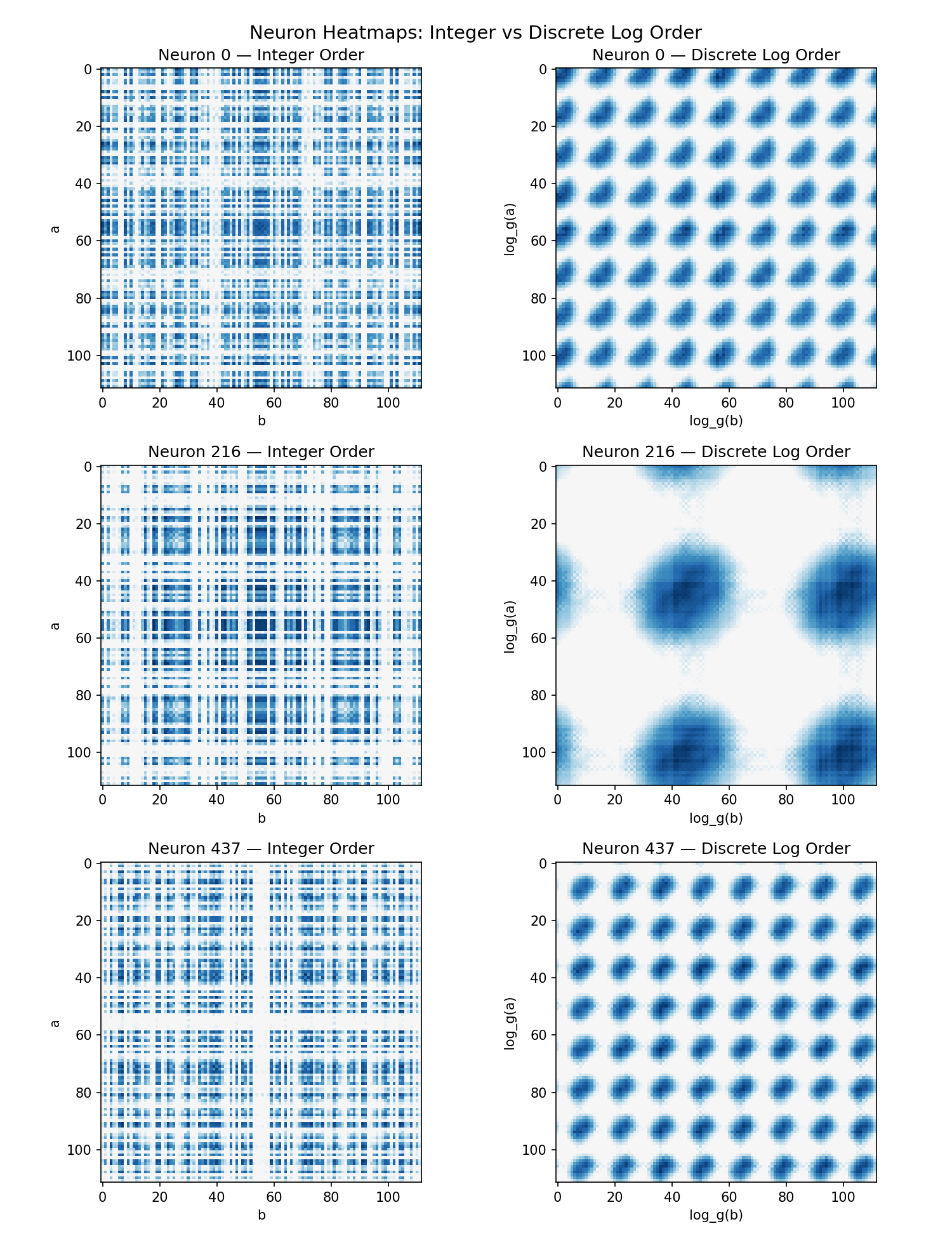}
\caption{Neuron heatmaps in integer order (left) vs.\ discrete-log order (right).
In log-order, diagonal stripes emerge, indicating periodicity in $\log_g(a) + \log_g(b)$.}
\label{fig:heatmaps}
\end{figure}

When neuron activation heatmaps are reordered by discrete logarithm, clear diagonal-stripe patterns emerge (Figure~\ref{fig:heatmaps}).
These stripes indicate the neuron computes a periodic function of $\alpha + \beta$, which is the signature of the trigonometric identity $\cos(k(\alpha + \beta))$ operating in log-space: the same mechanism Nanda et al.\ found for addition, but applied after the discrete-log transformation.

\subsection{SVD in Log-Order}

\begin{figure}[t]
\centering
\includegraphics[width=\columnwidth]{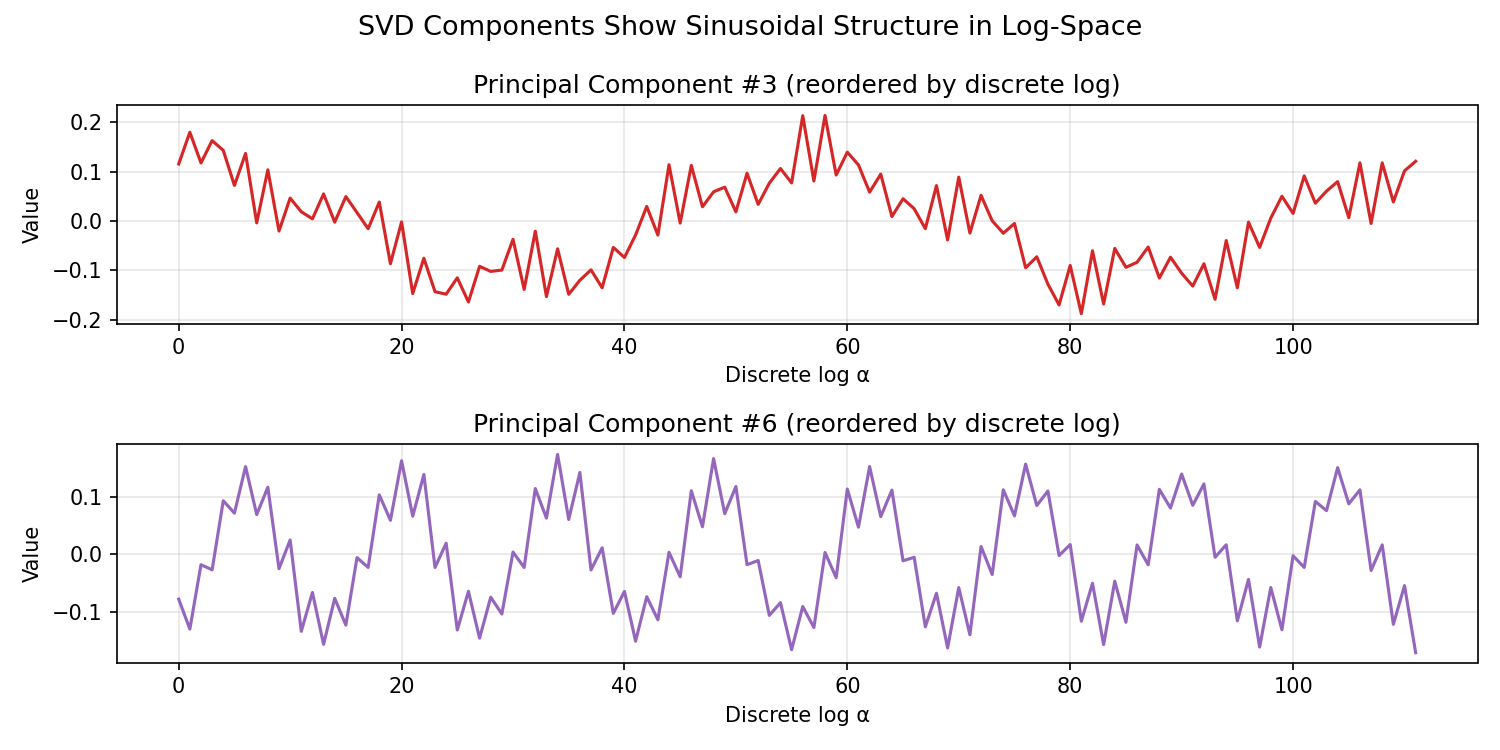}
\caption{Principal components of $W_E$ reordered by discrete logarithm show sinusoidal structure, confirming the embedding encodes multiplicative characters.}
\label{fig:svd}
\end{figure}

We perform SVD on $W_E$ and reorder principal components by discrete logarithm.
In integer order, all components appear noisy.
In log-order, several components reveal sinusoidal structure (Figure~\ref{fig:svd}), consistent with the embedding encoding multiplicative characters.
The effective rank is 10.3 (out of 112), confirming the embedding lives in a low-dimensional subspace.

\subsection{Discussion}

\textbf{The Discrete-Log Clock algorithm.}
Our results show the transformer solves modular multiplication by:
(1) embedding each integer $a$ as sinusoidal functions of $\log_g(a)$;
(2) routing embeddings via attention to the output position;
(3) applying trig identities in the MLP to compute $\cos(k(\alpha + \beta)/q)$;
(4) scoring each candidate $c$ by alignment with $\cos(k\, \log_g(c)/q)$, where only the correct answer $c = ab \bmod p$ achieves constructive interference across all frequencies.

\textbf{Why prior work missed this.}
\citet{furuta2024interpreting} and \citet{doshi2024grokking} analyzed with the additive DFT.
A function periodic in $\log_g(a)$ appears non-periodic in $a$ because the discrete logarithm is a nonlinear permutation of the integers.
The ``all-frequencies'' finding is correct in the additive basis but misleading because it conflates basis mismatch with algorithmic complexity.

\textbf{Overfitting and generalization.}
Grokking is itself an overfitting phenomenon: the model achieves zero training loss by epoch 500 (memorization), then continues training for thousands of epochs before discovering the generalizing algorithm.
Weight decay penalizes the large, unstructured weights needed for memorization, gradually favoring the compact Fourier representation that generalizes.
We do not use cross-validation because the task is deterministic (no label noise); the 70\% held-out test set directly measures generalization.
The model achieves 100\% test accuracy after grokking, confirming complete generalization.

\textbf{Limitations.}
We do not prove this is the only possible algorithm; prior work on addition \citep{zhong2023clock} shows different architectures can yield different solutions.

\section{Conclusion and Future Work}

We have shown that a transformer trained on modular multiplication learns the ``Discrete-Log Clock'': it reduces multiplication to addition in discrete-log space, then applies the same Fourier/trigonometric mechanism known from addition.
The key methodological insight is that the analysis basis must match the algebraic structure of the task.

In follow-up experiments, we confirmed the Discrete-Log Clock across 10 primes ($p = 59$ to $113$), all exhibiting sparse multiplicative spectra with 4 to 7 key frequencies.
We also trained across multiple random seeds on $p = 113$ and found the mechanism is universal ($\sim$80\% of seeds), with consistent sparsity (Gini 0.45--0.61) despite varying key frequencies.
We have also achieved preliminary results on modular exponentiation ($a^b \bmod p$) with a 2-layer transformer that achieves 100\% accuracy on $a^b \bmod 41$, to our knowledge the first demonstration of a transformer perfectly learning this operation.
The mechanism for exponentiation remains under active investigation.

The ``right basis'' principle extends naturally: for any algebraic task, the Fourier transform on the relevant group should reveal interpretable structure.
Future work includes extending this methodology to polynomial operations, and multi-layer architectures.

\section*{Appendix}

\subsection*{A.0\quad Reproducibility}

Code is available at: \url{https://github.com/danny-cpp/Discrete-Log-Clock}

\subsection*{A.1\quad The Discrete-Log Clock Derivation}
\label{app:clock}

This appendix derives the algorithm the transformer approximately implements for modular multiplication.
Let $\omega_k = 2\pi k / q$.

\vspace{0.3em}
\begin{tabular}{ll}
\toprule
Symbol & Meaning \\
\midrule
$g = 3$ & Primitive root of $p = 113$ \\
$\alpha, \beta, \gamma$ & Discrete logs of $a$, $b$, candidate $c$ \\
$\mathcal{K} = \{2,8,47,56\}$ & Key frequencies \\
$\gamma^* = (\alpha{+}\beta) \bmod q$ & Correct answer in log-space \\
\bottomrule
\end{tabular}

\subsection*{A.2\quad The Embedding}

Each dimension $j$ of $W_E$, viewed as a function of $\alpha$ across all 112 elements, is approximately a sinusoid at a key frequency $k_j \in \mathcal{K}$:
\begin{equation}
W_E[g^\alpha,\; j] \;\approx\; A_j \sin(\omega_{k_j} \alpha + \phi_j)
\label{eq:embed_dim}
\end{equation}
A single row $W_E[a]$ has no visible structure: it is a point-sample of all 128 waves evaluated at $\alpha = \log_g(a)$, plus noise.

\subsection*{A.3\quad Attention}

Attention routes information from positions $a$ and $b$ to position $=$\,.
After attention, the residual stream at $=$ approximately encodes the Fourier components of both inputs at the key frequencies.

\subsection*{A.4\quad The MLP (Trigonometric Addition)}

Writing $c_k(x) = \cos(\omega_k x)$ and $s_k(x) = \sin(\omega_k x)$ for brevity, the attention and MLP layers together approximately implement:
\begin{align}
c_k(\alpha{+}\beta) &= c_k(\alpha)\,c_k(\beta) - s_k(\alpha)\,s_k(\beta) \label{eq:trig_cos} \\
s_k(\alpha{+}\beta) &= s_k(\alpha)\,c_k(\beta) + c_k(\alpha)\,s_k(\beta) \label{eq:trig_sin}
\end{align}
After the MLP, the residual stream approximately encodes $c_k(\alpha{+}\beta)$ and $s_k(\alpha{+}\beta)$ for each $k \in \mathcal{K}$.

\subsection*{A.5\quad Scoring (Dot Product with Unembedding)}

The unembedding column for candidate $c$ approximately encodes $(\cos(\omega_k\gamma),\, \sin(\omega_k\gamma))$ for each $k$.
The logit for candidate $c$ is the dot product of the final residual with this column.
Since $\cos A \cos B + \sin A \sin B = \cos(A - B)$, each frequency's contribution reduces to:
\begin{equation}
\boxed{\;\text{logit}(c) \;\approx\; \sum_{k \in \mathcal{K}} A_k \cos\!\left(\omega_k(\alpha + \beta - \gamma)\right)\;}
\label{eq:score}
\end{equation}

\subsection*{A.6\quad Constructive Interference}

\textbf{Correct answer} ($\gamma^* = (\alpha{+}\beta) \bmod q$):
$$\text{logit}(c^*) \approx \sum_{k} A_k \cos(0) = \sum_{k} A_k \quad \text{(maximum)}$$
All cosines equal 1; all frequencies constructively interfere.

\textbf{Wrong answer} ($\Delta = \alpha{+}\beta{-}\gamma \neq 0$):
$$\text{logit}(c) \approx \sum_{k} A_k \cos(\omega_k \Delta) < \sum_{k} A_k$$
Cosines at different frequencies take values in $[-1,1]$, partially canceling.
With 4 incommensurate frequencies, destructive interference ensures no wrong answer approaches the correct answer's score.

\bibliography{references}

@inproceedings{power2022grokking,
  title={Grokking: Generalization Beyond Overfitting on Small Algorithmic Datasets},
  author={Power, Alethea and Burda, Yuri and Edwards, Harri and Babuschkin, Igor and Misra, Vedant},
  booktitle={ICLR 2022 Workshop on MATH-AI},
  year={2022}
}

@inproceedings{nanda2023progress,
  title={Progress Measures for Grokking via Mechanistic Interpretability},
  author={Nanda, Neel and Chan, Lawrence and Lieberum, Tom and Smith, Jess and Steinhardt, Jacob},
  booktitle={International Conference on Learning Representations},
  year={2023}
}

@inproceedings{zhong2023clock,
  title={The Clock and the Pizza: Two Stories in Mechanistic Explanation of Neural Networks},
  author={Zhong, Ziqian and Liu, Ziming and Tegmark, Max and Andreas, Jacob},
  booktitle={Advances in Neural Information Processing Systems},
  year={2023}
}

@article{doshi2024grokking,
  title={Grokking Modular Polynomials},
  author={Doshi, Darshil and He, Tianyu and Das, Aritra and Gromov, Andrey},
  journal={arXiv preprint arXiv:2406.03495},
  year={2024}
}

@article{furuta2024interpreting,
  title={Towards Empirical Interpretation of Internal Circuits and Properties in Grokked Transformers on Modular Polynomials},
  author={Furuta, Hiroki and Minegishi, Gouki and Iwasawa, Yusuke and Matsuo, Yutaka},
  journal={Transactions on Machine Learning Research},
  year={2024}
}

@inproceedings{chughtai2023toy,
  title={A Toy Model of Universality: Reverse Engineering How Networks Learn Group Operations},
  author={Chughtai, Bilal and Chan, Lawrence and Nanda, Neel},
  booktitle={International Conference on Machine Learning},
  year={2023}
}

@inproceedings{mccracken2025universal,
  title={Uncovering a Universal Abstract Algorithm for Modular Addition in Neural Networks},
  author={McCracken, Gavin and Moisescu-Pareja, Gabriela and Letourneau, Vincent and Precup, Doina and Love, Jonathan},
  booktitle={Advances in Neural Information Processing Systems},
  year={2025}
}

@inproceedings{stander2024grokking,
  title={Grokking Group Multiplication with Cosets},
  author={Stander, Dashiell and Yu, Qinan and Fan, Honglu and Biderman, Stella},
  booktitle={International Conference on Machine Learning},
  year={2024}
}
\bibliographystyle{icml2026}

\end{document}